\documentclass[conference]{IEEEtran}
\usepackage{times}

\usepackage[numbers]{natbib}
\usepackage{multicol}
\usepackage[bookmarks=true]{hyperref}
\usepackage{graphicx}
\usepackage{caption}
\usepackage{amsmath,amssymb}
\usepackage{stackengine}
\usepackage{xcolor}
\usepackage{algorithm}
\usepackage{algpseudocode}
\usepackage{algorithmicx}

\algdef{SE}[SUBALG]{Indent}{EndIndent}{}{\algorithmicend\ }%
\algtext*{Indent}
\algtext*{EndIndent}

\begin{document}

% paper title
\title{DinerDash Gym: A Benchmark for Policy Learning in High-Dimensional Action Space}

\author{\authorblockN{Siwei Chen, Xiao Ma, David Hsu}
\authorblockA{National University of Singapore\\
\{siwei-15, xiao-ma, dyhsu\}@comp.nus.edu.sg}}

\maketitle

% 1. problem
% 2. problem statement
% 3. what to do
% 4. Biggest take away
% 5. conclusion

\begin{abstract}
It has been arduous to assess the progress of a policy learning algorithm in the domain of hierarchical task with high dimensional action space due to the lack of a commonly accepted benchmark. In this work, we propose a new light-weight benchmark task called Diner Dash for evaluating the performance in a complicated task with high dimensional action space. In contrast to the traditional Atari games that only have a flat structure of goals and very few actions, the proposed benchmark task has a hierarchical task structure and size of 57 for the action space and hence can facilitate the development of policy learning in complicated tasks. On top of that, we introduce Decomposed Policy Graph Modelling (DPGM), an algorithm that combines both graph modelling and deep learning to allow explicit domain knowledge embedding and achieves significant improvement comparing to the baseline. In the experiments, we have shown the effectiveness of the domain knowledge injection via a specially designed imitation algorithm as well as results of other popular algorithms. The code is available online \footnote{https://github.com/AdaCompNUS/diner-dash-simulator}.
\end{abstract}

\IEEEpeerreviewmaketitle

\section{Introduction}
% Projection writing:

% 1. what to do

%     a) IL is good, xxx
    
%     b) problem of current use of IL
    
%     c) what to do

% 2. why and the benefits

%     a) decomposition simplifies the task
    
%     b) help to address sample efficiency, effective
    
%     c) may improve the upper bound given a imperfect training set

% 3. rough idea and steps

Imitation learning (IL) and reinforcement learning (RL) have shown remarkable success in completing complex and challenging tasks \cite{finn2016guided, schaal1999imitation, muller2006off, bojarski2016end, mahler2017learning, wang2019monocular}. The most representative algorithms such as Behaviour Cloning (BC) \cite{pomerleau1989alvinn} and Generative Adversarial Imitation Learning (GAIL) is widely studied and used in many works \cite{ho2016generative,fu2017learning, pytorchrl}.
Real-world decision-making problems often have a high-dimensional and highly structured action space, e.g., traffic light control with thousands of actions \cite{zahavy2018learn}.
However, both imitation learning and reinforcement learning algorithms do not always generalize to tasks with high dimensional state space and high dimensional action space. For example, the maximum entropy inverse reinforcement learning (MaxEnt IRL) \cite{ziebart2008maximum}, which matches the expert trajectories by shaping a reward function, works well in a small state space task but not a high dimensional state space task. 

% In this work, we propose a new benchmark task, which has high dimensional state space, high dimensional action space as well as the hierarchical structure, to quantify the progress of newly proposed learning algorithms.

% advantages:
% - a step towards the real problems, with negligible cost.
% - provide a simple algo to show how domain knowledge can help

% Similar problems identified in \cite{dulac2015deep,zahavy2018learn}, real tasks like controlling power grid and traffic lights with thousands of actions have significant impacts to the society. How can we train a policy that can solve high dimensional action space task remains a key challenge. Moreover, the cost to set up such complex training environment is also a reasonable concern. The proposed benchmark task, Diner Dash, moves one step further to provide a relatively complex training environment with negligible training cost to run such simulator. Furthermore, for people who want to study how existing domain knowledge can affect the learning performance, we do provide a benchmark score done by a simple imitation learning algorithms with specially structured policy representation to allow explicit domain knowledge injection. The detail of this policy can be found in section \ref{sec:method}.

Efficient simulation for evaluating IL / RL algorithms in high-dimensional action space is, however, hard to acquire. Most of the existing benchmarks have a simple action space, e.g., Atari games with up to 6 actions and Mujoco \cite{duan2016benchmarking} with less than 20 dimensions of actions. Recent works proposed solutions to specific tasks with high-dimensional action spaces, e.g., real-world YouTube recommendation system \cite{covington2016deep} and StartCraft Learning Environment \cite{vinyals2017starcraft}. However, the simulation cost of these tasks is very high, and the training difficulty is not only about the action space itself but also with the complexity of the games.

% By making use of the benchmark task, people can gain advantages in the following three ways. Firstly, it is easy to quantify the progress by comparing to other popular imitation algorithms such as Behaviour cloning and GAIL. Secondly, a heuristic expert policy is provided for both expert trajectories generation as well as the estimation of the expert performance. Since the expert policy is fixed for all the algorithms that use this benchmark task, it is fair to compare with each other without re-training the expert policy by self-implemented RL algorithms. Using self-implemented RL algorithms as an expert result in inconsistent expert policy cross different works and hence non-comparable numerical results. Thirdly, for people who want to study how existing domain knowledge can affect the learning performance, we do provide a benchmark score done by a simple imitation learning algorithms with specially structured policy representation to allow explicit domain knowledge injection. The detail of this policy can be found in section \ref{sec:method}.

In this work, we proposed a benchmark task called Diner Dash for high dimensional action space with hierarchical structure in the Open-AI Gym environment \cite{brockman2016openai}. The benchmark task is light-weight and fast to run in a gym environment. Existing training environments like StarCraft is too complicated and is resource-consuming to train. Other existing benchmarks like Atari games are light-weight, however, cannot provide a high dimensional action space environment. DinerDash is hence a suitable and balanced benchmark task for evaluating policy in a task with high dimensional action space.
Furthermore, we have implemented a simple imitation learning algorithm DPGM to study how existing domain knowledge can help with the performance. DPGM leverages on the decomposition and models each decomposed task as a factor graph. Following this approach, DPGM significantly beat the naively implemented baselines such as behaviour cloning (BC). Finally, We compare with popular imitation learning algorithms such as Behaviour cloning and GAIL as well as popular RL methods such as PPO \cite{schulman2017proximal} with their results reported in the experiments. 

\begin{figure}[h]
  \centering
  \includegraphics[width=0.9\linewidth]{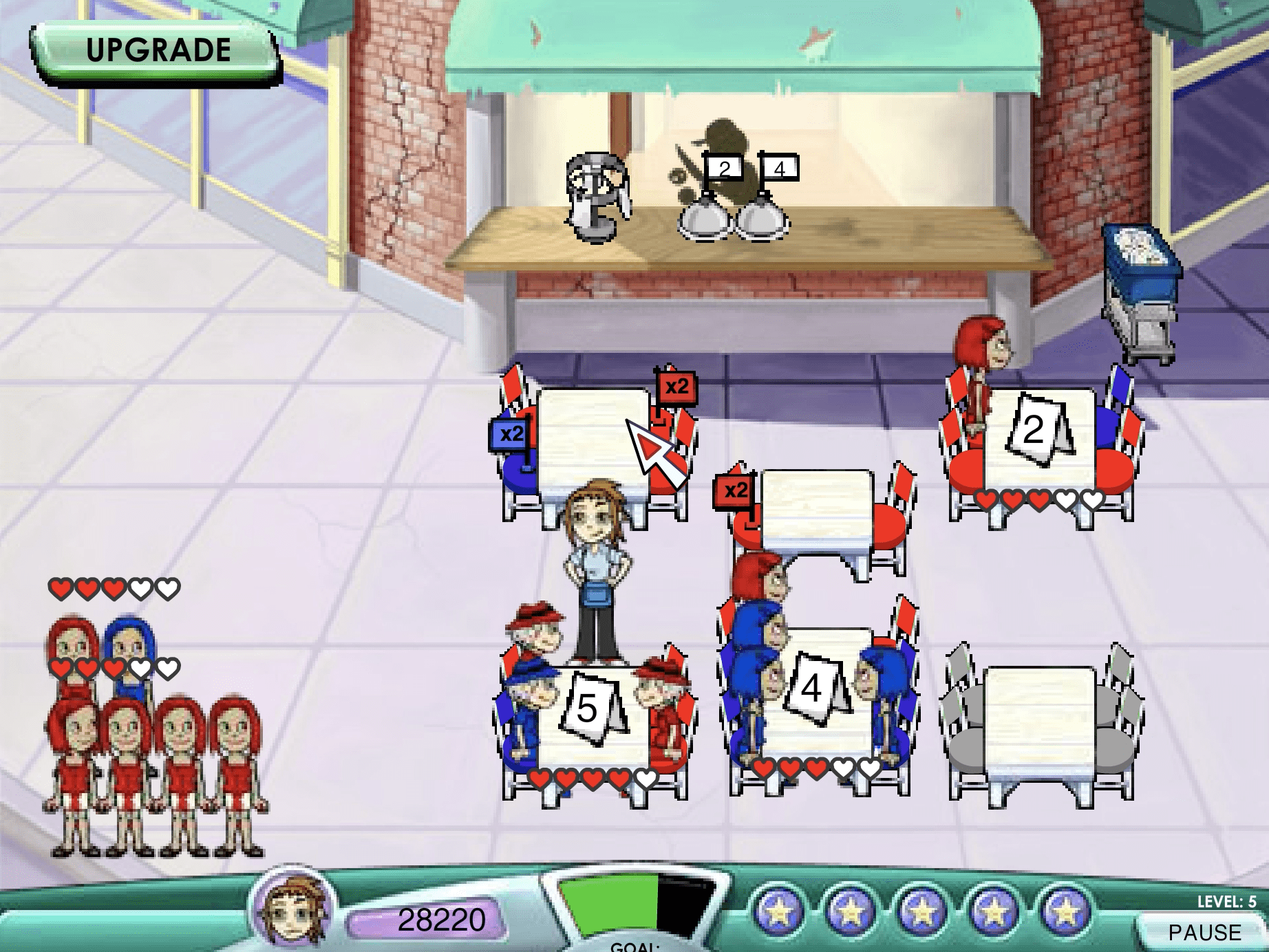}
  \caption{Diner Dash Simulator. The task is called Diner Dash, where the player runs a restaurant and serves the customer as many as possible. The difficulty of this task is overwhelming customers and a long planning horizon. Each group of customers takes multiple steps to serve and have to serve multiple customers at the same time. The agent has to make the right choice for every step; otherwise, the customer will run away and brings in large negative rewards.}
  \label{Fig:diner_dash}
\end{figure}

\section{Related Work}
There are many benchmark works available, and some of them target on specific problems. \cite{dejong1994swinging} introduced Acrobot and \cite{moore1990efficient} brought out mountain car model. Modern benchmark works including RL-Toolbox, Beliefbox, RLLib and RLLab \cite{duan2016benchmarking,neumann2006reinforcement, kochenderferjrlf, abeyruwanrllib} provide a good training platform with tasks ranging from Cart-Pole Balancing, Mountain Car, Atari games to locomotion tasks and partially observable tasks.

Other benchmark works have also focused on the high dimensional action space. For example, \cite{schaul2010pybrain} uses a 16-DOF humanoid robot, \cite{yamaguchi2010skyai} introduces 17-DOF humanoid robot task for crawling and \cite{dann2014policy} introduces a 20-link pole balancing task. Other tasks such the RoboCup Keepaway \cite{stone2005keepaway} introduces a multi-agent task which has high dimensional actions. However, most of the benchmark tasks above have a relatively small action space comparing to the real problem, which has thousands of actions such as traffic light control. The proposed benchmark task Diner Dash works on this to provide a more realistic training environment with high dimensional action space to quantify the process of a learning algorithm.

\section{Task Description}

The game in Figure \ref{Fig:diner_dash} called Diner Dash is proposed as the benchmark task. The task has high dimensional state space, 40 dimensions to be exact, and 57 actions for the action space. The player is running a restaurant by controlling a waitress to serve customers as many as possible. As shown in the picture, the restaurant has six tables with different sizes and up to 7 waiting groups, on the left side and with different sizes, to be served. For each group of people, the player needs to allocate a table for them, collect orders, submit orders, pick up food, serve food, collect bills, clean table and finally return the dish to the dish collection point.  There is a happiness value of each group of people, represented in the form of hearts, and the happiness value will decrease if they wait too long. Once the happiness value reaches zero, the customer runs away, and the player loses one life. There is a maximum of 5 lives of each player, and the game ends when the player loses five groups of customers. 

Given all the properties above of the task, Diner Dash is a challenging task, with high dimensional action space, high dimensional state space, infinite horizon, hierarchical structure and requires sub-tasks to be completed in parallel. Such a tough task gives a better training environment which is closer to the real-world problems, for example, traffic light control, comparing to typical RL benchmark tasks such as Atari games.

% \section{Algorithms}
% \subsection{Behaviour Cloning (BC)}

% \subsection{GAIL}

% \subsection{PPO}

% \section{Algorithms}
% This sections explains what algorithms are evaluated in the benchmark task and why they are chosen. In the domain of imitation learning, there are mainly two trends: state-action matching and trajectory matching. The state-action matching refers to behaviour cloning (BC) and its variants. While the most representative methods in trajectory matching is GAIL \cite{ho2016generative}, which uses a GAN like structure to train a policy to generate trajectories like expert. Inverse reinforcement learning (IRL) approach is special case of trajectory matching, which imitate the expert by learning a reward function. However, the most representative approach MaxEnt IRL \cite{ziebart2008maximum} does not work for high dimensional state space. In the reinforcement learning domain, PPO \cite{schulman2017proximal} is a good representative and shows a great success in many tasks.

% Therefore, BC, GAIL and PPO are selected in the evaluation of the benchmark task, representing the trends of state-action matching, trajectory matching and reinforcement learning. In this work, we used a naive implementation of BC, which has two fully connected layers and one dropout layer in between.

\section{Domain Knowledge Embedding}
\label{sec:method}

In this section, we proposed a simple imitation learning algorithm that combines both graph modelling and deep learning to allow explicit domain knowledge injection. The purpose of this algorithm is to provide a baseline for people who want to study how existing domain knowledge can affect the final performance of a policy.

\subsection{Overview}

\begin{figure}[h]
  \centering
  \includegraphics[width=0.7\linewidth]{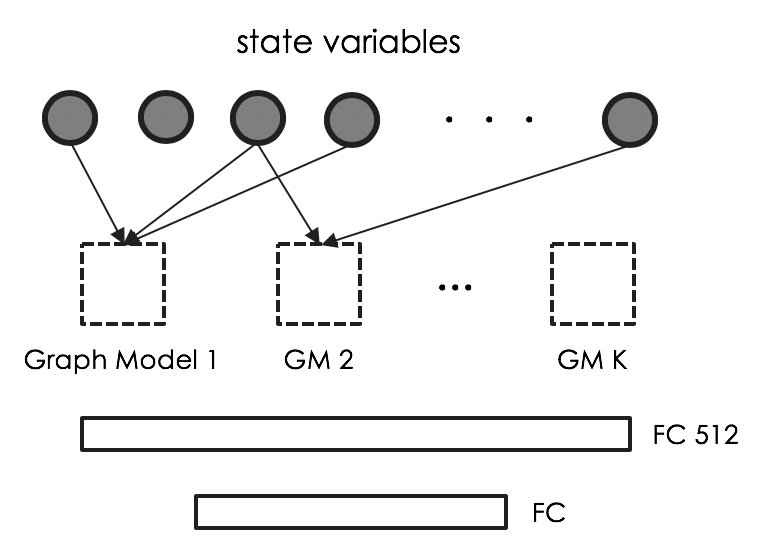}
  \caption{Decomposed Policy Graph Modeling (DPGM) Structure. The input is the current state vector, and the output is a vector of action values. Instead of learning the entire policy based on the complex state space, we can decompose the problem into K sub-problems to learn a simple policy for each particular action. Each small policy is modelled by a graph model in the form of the factor graph, which allows the expert to inject prior knowledge easily. After the graph modelling, we apply another two fully connected layers to re-weight the importance of each action with supervised learning.}
  \label{Fig:structure}
\end{figure}

In a closer view, our approach first decompose the entire state space into K small tasks and model each task by a factor graph, see figure \ref{Fig:structure}. More formally, the original task can be formulated with a standard Markov decision process (MDP) $ M = \{ S, A, T, R, \gamma \}$ with state space $S$, action space $A$, state transition probability distribution $T$, reward function $R$, and discount factor $\gamma$. After decomposition, it becomes $ M = \{ \{S_1,S_2, ..., S_K\}, A, T, R, \gamma \}$ with $S_k \subseteq S$. 

Our goal is to train the K graph models $g_{\theta_k}(s_k)$ to imitate the expert on each action and output the final re-weighted result by the fully connected layers $f_\alpha$, see equation \ref{eq:goal}.

\begin{align}
    \alpha^* , \theta^* = arg \min_{\alpha, \theta} \sum_{(s,a) \in D} L(a, \: \: f_\alpha(g_{\theta_1}(s_1), ... , g_{\theta_k}(s_k)))
    \label{eq:goal}
\end{align}

\subsection{Task Decomposition Along Action Space}

\begin{figure}[h]
  \centering
  \includegraphics[width=0.8\linewidth]{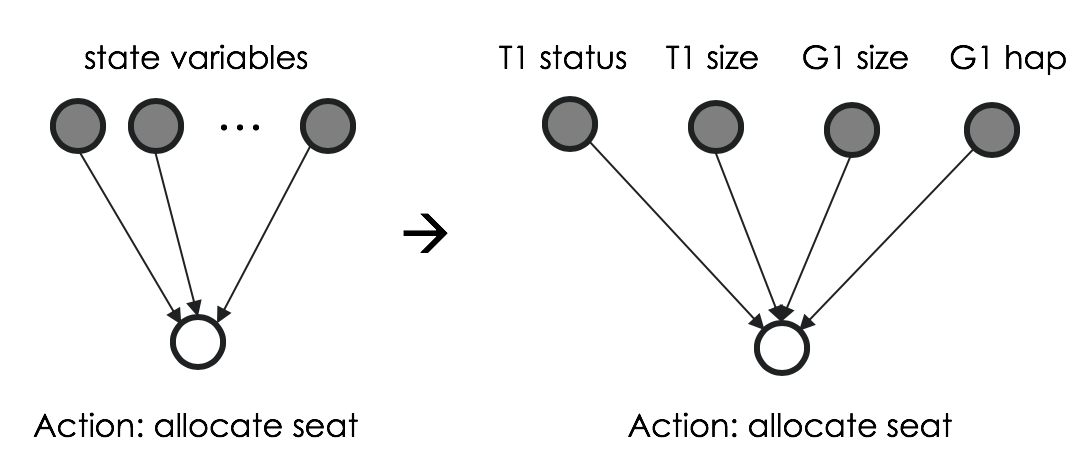}
  \caption{Task Decomposition Along with Action Space.}
%   Although, a task may have a high dimensional state space, the actual related state for each action can be very small. The above shows an example, in a restaurant daily operation, the action to allocate a seat for a waiting customer group to a particular table may only need to consider whether the table is ready and whether the table size fits the customers. Other table status are not relevant to this particular action.
  \label{Fig:decomposition}
\end{figure}

% The expert should not only provide demonstrations but also explain what key features the agent should pay attention to. To better explain the idea, we take the example of the Diner Dash task, which is used as the experiment task in this work.

Figure \ref{Fig:decomposition} is an example of task decomposition along with each action, wherein this task, the agent needs to allocate a table for a group of people. Even though the full state space has 40 variables, the key factors the agent needs to pay attention are only 4 variables: the table status, the table size, the group size and the group happiness. Other irrelevant state variables are discarded, leaving a much lesser state space. However, the task decomposition introduces an underlining assumption that the new sub-state space should be sufficient for decision making. 

\textbf{Assumption 1} The decomposed state space $S_k$ of action is sufficient for decision making policy $\pi_k(s_k)$, where $\pi_k$ only has one action and learns that action from the expert demonstration.

% Following this approach, the original task now can be decomposed into several small tasks along with each action. Each small task has only one action output. Such decomposition process reduces the state space heavily from $O(M^{40})$ to about $O(M^4)$, where $M$ is the average number of state values for each dimension.  

\subsection{Decomposed Policy Graph Modeling (DPGM)}
After decomposition, it is necessary to reason out why the expert chooses the action by modelling the action with a factor graph. 

\begin{figure}[h]
  \centering
  \includegraphics[width=0.8\linewidth]{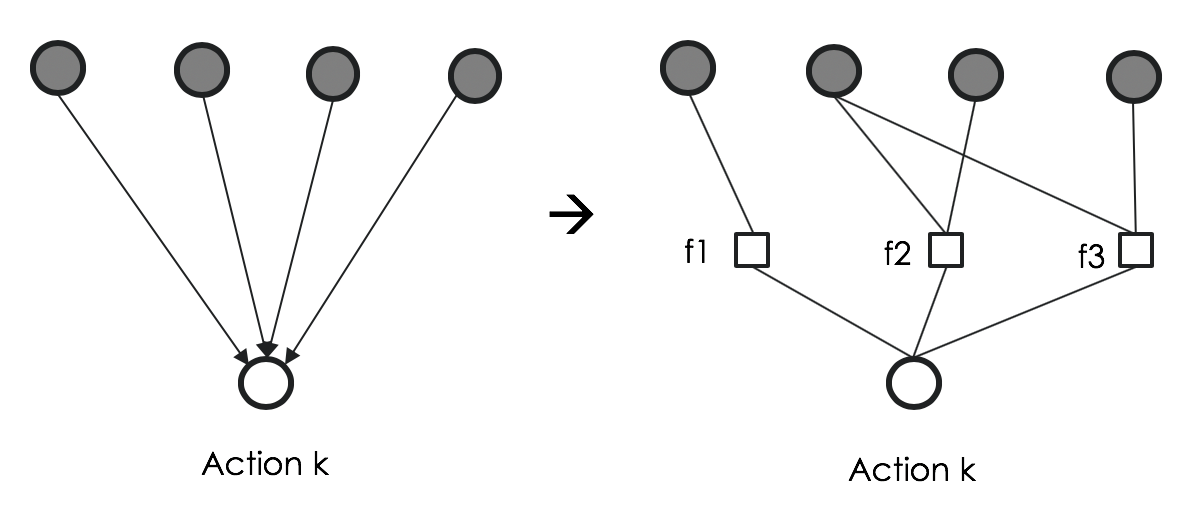}
  \caption{Policy Graph Modeling as a Factor Graph.}
  \label{Fig:PGM}
\end{figure}
Figure \ref{Fig:PGM} is an example of modelling the correlations between the input variable and the action output. In this example, the action output is modelled by three factors shown on the right side. By modelling with the factor graph, we can have lesser parameters to train and hence is more data-efficient. The structure of the factor graph is provided by the expert who has the domain knowledge, understanding the correlations between input variables and the probability to choose this action. If the internal correlation groups are not clear to the expert, we can always use one big factor or a neural network to model the distribution. 

Moreover, this factor graph structure allows the expert to group input states as factors and allows the expert to specify some known correlations. For example, the factor $f_1$ can be modelled as a correlation between table happiness and the the action whether to serve that table as the expert knows the lower the happiness of the table the higher priority to serve that table.

The joint distribution of the factor graph is shown below:
\begin{align}
    p(\pmb{x},y) &= \frac{1}{Z} \prod_i \varphi_i(\pmb{x_i},y)\\
    Z &= \sum_{\pmb{x},y} \prod_i \varphi_i(\pmb{x_i},y) \nonumber
\end{align}
$\varphi_i$ is the ith potential function corresponding to factor $f_i$ in the factor graph. $S_k$ is the mapped sub-state space and $\pmb{x_i} \subseteq S_k$ are the input variables required by factor $\varphi_i$. Therefore, the joint distribution is the product of all the factors and then normalized by the partition function Z. Given the joint distribution, the goal of the graph model is to model how the expert selects this action. The inference equation is shown below.
\begin{align*}
	p_k(\pmb{x}) &\equiv p(y=1|\overline{\pmb{x}}) \\
	&= \frac{p(\overline{\pmb{x}},y=1)}{\sum_y p(\overline{\pmb{x}},y)} \\ 
	 &= \frac{\frac{1}{Z} \prod_i \varphi_i(\overline{\pmb{x_i}},y=1)}{\sum_y \frac{1}{Z} \prod_i \varphi_i(\overline{\pmb{x_i}},y)} \\
	 &= \frac{\prod_i \varphi_i(\overline{\pmb{x_i}},y=1)}{\prod_i \varphi_i(\overline{\pmb{x_i}},y=0) + \prod_i \varphi_i(\overline{\pmb{x_i}},y=1)}
\end{align*}
$\overline{\pmb{x}}$ means observed input variables. The agent selects actions based on the soft-max of each state-action values. Since the scale of each action is different, it is hard to choose the action based on the absolute values. Therefore, another layer of fully connected layers will be added after the graph modelling layers to re-weight the importance of each action. 

\begin{algorithm}
    \caption{Decomposed Policy Function Modeling}
    \begin{algorithmic}[1]
        \Require Domain Knowledge and full state $S$
        
        \State Gather demonstrations $D = \{s_1,a_1...,s_t,a_t\}$
        \State Map each state $s_t \to s_{k,t}$ by domain knowledge
        \State Save mapped data as $D_{k}$
        
        \For{each action $k$}
        \State Factorize and group variables into a factor graph $\mathcal{G}$
        \State Joint distribution to be $p(\pmb{x},y) = \frac{1}{Z} \prod_i \varphi_i(\pmb{x_i},y)$
        \State Update $\min_{\theta} L = \sum_{\pmb{x},y \in D_{a_i}} (\: p(y=1|\pmb{\overline{x_t}}) - y_t)^2$
        \EndFor
        \State Train the re-weighting layer with the expert action.
        \State Update the fully connect layers $f_\alpha$

    \end{algorithmic}
        \label{alg:DPGM}
\end{algorithm}

\subsection{Training Method}

For each action, with the graph model, we can update the graph parameters by the equation \ref{Eq:update}. The training data comes from the demonstration and is in the form of $(\overline{\pmb{x_t}},y_t)$, where $\overline{\pmb{x_t}}$ is the observed sub-state variables and $y_t$ is a binary value whether the expert selects the action. For example, there are actions A and B, sub-state space $\pmb{x_A}$ and $\pmb{x_B}$. If the expert selects A at time step t, then $(\overline{\pmb{x_{A,t}}},1)$ is the training data for action A and $(\overline{\pmb{x_{B,t}}},0)$ for action B.
\begin{align}
    \min_{\theta} L = \sum_t (\: p(y=1|\overline{\pmb{x_t}}) - y_t)^2
    \label{Eq:update}
\end{align}

% The overall network structure is shown in figure \ref{Fig:structure}. Instead of learning the entire policy based on the complex state space, we can decompose the problem into K sub-problems to learn a simple policy for each particular action with a much smaller state space. Each small policy is modelled by a graph model in the form of the factor graph, which allows the expert to inject prior knowledge easily. After the graph modelling, we apply another two fully connected layers to re-weight the importance of each action with supervised learning. 

The fully connected re-weighting layer is trained with the supervision of the expert demonstrations,  $(\pmb{x},y)$. We perform an end-to-end training, including the graph policy models and the re-weighting layers with the cross-entropy loss.

\section{Experiment}

\subsection{Collecting Demonstration Data}
Collecting demonstration data is a time consuming and even costly process at some times. To address the issue, we used a heuristic policy function together to be the expert for demonstration data collection. The simulator wrapped by Gym Environment is used to collect the demonstration data. In this experiment, a total of 274 trajectories with 163120 state-action pairs, are collected for the expert demonstration.

\subsection{Algorithms}
% This sections explains what algorithms are evaluated in the benchmark task and why they are chosen. In the domain of imitation learning, there are mainly two trends: state-action matching and trajectory matching. The state-action matching refers to behaviour cloning (BC) and its variants. While the most representative methods in trajectory matching is GAIL \cite{ho2016generative}, which uses a GAN like structure to train a policy to generate trajectories like expert. Inverse reinforcement learning (IRL) approach is special case of trajectory matching, which imitate the expert by learning a reward function. However, the most representative approach MaxEnt IRL \cite{ziebart2008maximum} does not work for high dimensional state space. In the reinforcement learning domain, PPO \cite{schulman2017proximal} is a good representative and shows a great success in many tasks.

% Therefore, BC, GAIL and PPO are selected in the evaluation of the benchmark task, representing the trends of state-action matching, trajectory matching and reinforcement learning.

We compare the proposed DPGM with two commonly used imitation learning algorithms on DinerDash, standard behaviour cloning (BC) \cite{pomerleau1989alvinn} and one of the SOTA methods, GAIL~\cite{ho2016generative} which is the representative of trajectory matching approaches in IL. We also evaluate the SOTA on-policy reinforcement learning method, PPO~\cite{schulman2017proximal, pytorchrl} and demonstrate the challenges standard RL algorithms would face in DinerDash.

For the standard BC, we used a network with two fully connected layers and one dropout layer in between. We directly predict the actions and use cross-entropy loss for training. For a fair comparison, both the propose DPGM and BC use the same set of expert demonstration data.

\subsection{DPGM Training Pipeline}
The detailed algorithm can be found in Algorithm \ref{alg:DPGM}. Firstly, the expert decomposes the task and selects the relevant states regarding each action. This can be done given assumption 1, where the teacher knows well of the task and is able to decide a sub-state space which is sufficient for decision making. Each action is then modelled by a factor graph where the expert can easily inject the prior knowledge.

In this experiment, action 1 to 6 corresponds to moving tables 1 to 6 and action 15 to 57 corresponding to allocating group x to table y are modelled by graph models. The other actions adopt the memorizing approach, which is to memorize the expert’s actions. This is because the sub-state space of the other actions is small, and all the choices can be easily memorized. The graph model parameters are learned from the demonstration data based on equation \ref{Eq:update}. 

\begin{figure}[h]
  \centering
  \includegraphics[width=1\linewidth]{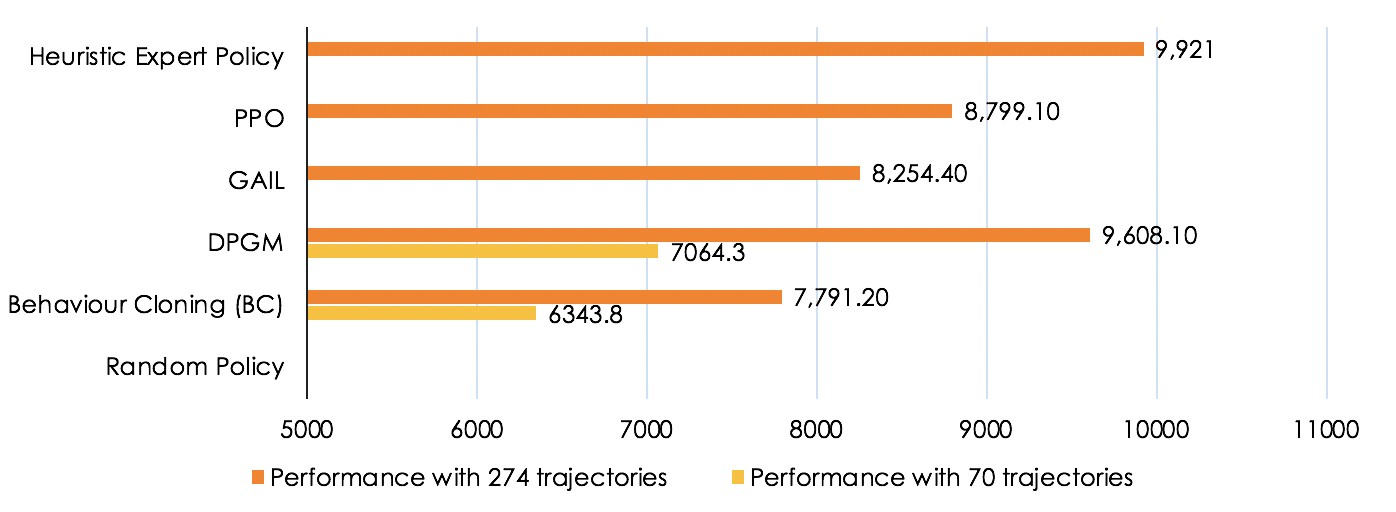}
  \caption{Final Performance in the Diner Dash Simulator. The top row is the heuristic policy which is also the expert that generates the training data. The second row is the proposed approach, named decomposed policy graph modelling (DPGM). The last row is a random agent and the score of the random agent is -1245.8, which cannot be displayed in the figure.  }
  \label{Fig:result1}
\end{figure}
\subsection{Results}
The experiment results can be found in Figure \ref{Fig:result1}. To truly identify the performance of all the agents, the difficulty of Diner Dash is adjusted to a higher level comparing to the original version. Total of 274 expert trajectories from the heuristic policy is collected for demonstration and the performance is the average score based on 100 episodes. In this experiment, we are comparing with standard behaviour cloning, which is a common baseline in imitation learning. The result shows that DPGM can achieve near-optimal performance in a complicated task such as Diner Dash with 40 dimensions of the state space and 57 actions in the action space. 

The behaviour cloning baseline shows us precisely what is the performance without using any existing prior knowledge and purely relies on the demonstration data. Eventually, It fails to converge to a near-optimal performance due to the insufficient data. This also shows that DPGM has a higher data sampling efficiency by decomposition and factor graph modelling.

On top of that, we have anther comparison between BC and DPGM with only 70 trajectories to study the sample efficiency of the proposed method. The performance of DPGM is still higher than BC, and this is reasonable due to all the optimizations to make use of the prior knowledge.

The results of PPO suggests that even the SOTA of RL has some difficulties to converge to the optimal performance after $3 \times 10 ^ 7$ steps. It is hard for the unstable RL policies to discover the correlation that is crucial, sparse and long-delayed in high dimensional action space task like Diner Dash: a wrong action may result in the customer loss only after a certain number of steps and the game ends when losing 5 customers. Given the large action space, effective exploration in RL becomes extremely difficult and the performance of PPO is highly affected. 

% For example, for every customer lost, the agent will receive a sizeable negative reward. However, the game will end after 5 customers lost, resulting in a sparse signal. On top of that, the result of customer loss is due to the actions of not serving them along the trajectories many steps ago. The sizeable negative penalty reward tells the policy not to select the actions that are selected just now. Nonetheless, the problem is that there are so many other actions and trying and error become more difficult than tasks with small action space. 

% {\color{red} The RL baselines are normally unstable in practice~\cite{pardo2018time}. The task of Diner Dash, however, has a low error tolerance: a wrong decision may result in huge loss in future steps, especially in overwhelming situations. As a result, PPO baseline fails to perform well. 

% For better computation efficiency, PPO selects the first-order derivative approach and avoids the second-order derivative computation. This choice introduces a small number of bad decisions, but a good deal for speed optimization. However, Diner Dash requires policy to have low error tolerance, where a wrong decision may result in loss of table after many steps later, especially in overwhelming situations. \cite{pardo2018time} also suggests that handling time-awareness in PPO can significantly improve the performance in specific time-limited tasks.}

GAIL shows a higher sample efficiency than BC. However, without any domain knowledge of the challenging Diner Dash, fitting the expert policy is difficult, compared with DPGM. GAIL learns a discriminator that shapes the policy to behave like the expert using GAN, which is notorious for its data hunger. However, given a limited expert demonstration in our experiments, training an effective discriminator is hard. As a result, GAIL achieves worse performance compared to PPO trained with ground truth reward function and DPGM. 

% given limited expert demonstration in our experiments, 

% However, given limited demonstration data

% According to the results, GAIL shows a higher sample efficiency than Behaviour Cloning (BC), but it is still not adequate. Under the limited number of demonstration data, GAIL learns a discriminator that shapes the policy to behave like the expert. Nevertheless, it is hard to learn a perfect reward signal that is even better than the ground truth reward function. Consequently, the performance of GAIL drops, compared to the PPO trained under the ground truth reward function.

In summary, such a problem, with long-delayed sparse reward and high dimensional action space, poses a challenge to the RL / IL policy, making the RL / IL policies harder to converge.

% \section{Limitations}
% One of the major limitations of this approach is that the state space needs to have semantic meanings. Same as other decomposition approaches, such as MaxQ planning, DPGM relies on the expert's domain knowledge to decompose the task into small pieces for each action. In reality, not all the tasks have a semantic state space and therefore this approach does not apply to those tasks, for example the tasks with raw images as input.

\section{Conclusion and Future Work} 
% We proposed the approach called decomposed policy graph modelling (DPGM), which allows the expert to inject the existing domain knowledge into the training model. The experiment result also indicates the significance of the prior knowledge and effectiveness of our approach to inject prior knowledge. Moreover, our approach is able to work on complicated tasks such as Diner Dash and helps to boost up the final performance as well as the data sampling efficiency.

We introduce DinerDash, a challenging light-weight benchmark for IL / RL algorithms with high-dimensional action space. DinerDash poses significant challenges to state-of-the-art IL / RL algorithms. We also introduce DPGM that decomposes policy space using factor graph with expert domain knowledge and outperforms all baselines.

However, DPGM relies on expert domain knowledge and is hard to generalize to less-structured tasks. In addition, it requires the state space to have semantic meanings in order to do the decomposition. We leave it for future study.

% \section*{Acknowledgments}

%% Use plainnat to work nicely with natbib. 

\newpage
\bibliographystyle{plainnat}
\bibliography{references}

\begin{thebibliography}{26}
\providecommand{\natexlab}[1]{#1}
\providecommand{\url}[1]{\texttt{#1}}
\expandafter\ifx\csname urlstyle\endcsname\relax
  \providecommand{\doi}[1]{doi: #1}\else
  \providecommand{\doi}{doi: \begingroup \urlstyle{rm}\Url}\fi

\bibitem[Abeyruwan()]{abeyruwanrllib}
Saminda Abeyruwan.
\newblock Rllib: Lightweight standard and on/off policy reinforcement learning
  library (c++), 2013.
\newblock \emph{URL http://web. cs. miami. edu/home/saminda/rilib. html}.

\bibitem[Bojarski et~al.(2016)Bojarski, Del~Testa, Dworakowski, Firner, Flepp,
  Goyal, Jackel, Monfort, Muller, Zhang, et~al.]{bojarski2016end}
Mariusz Bojarski, Davide Del~Testa, Daniel Dworakowski, Bernhard Firner, Beat
  Flepp, Prasoon Goyal, Lawrence~D Jackel, Mathew Monfort, Urs Muller, Jiakai
  Zhang, et~al.
\newblock End to end learning for self-driving cars.
\newblock \emph{arXiv preprint arXiv:1604.07316}, 2016.

\bibitem[Brockman et~al.(2016)Brockman, Cheung, Pettersson, Schneider,
  Schulman, Tang, and Zaremba]{brockman2016openai}
Greg Brockman, Vicki Cheung, Ludwig Pettersson, Jonas Schneider, John Schulman,
  Jie Tang, and Wojciech Zaremba.
\newblock Openai gym.
\newblock \emph{arXiv preprint arXiv:1606.01540}, 2016.

\bibitem[Covington et~al.(2016)Covington, Adams, and Sargin]{covington2016deep}
Paul Covington, Jay Adams, and Emre Sargin.
\newblock Deep neural networks for youtube recommendations.
\newblock In \emph{Proceedings of the 10th ACM conference on recommender
  systems}, pages 191--198, 2016.

\bibitem[Dann et~al.(2014)Dann, Neumann, Peters, et~al.]{dann2014policy}
Christoph Dann, Gerhard Neumann, Jan Peters, et~al.
\newblock Policy evaluation with temporal differences: A survey and comparison.
\newblock \emph{Journal of Machine Learning Research}, 15:\penalty0 809--883,
  2014.

\bibitem[DeJong and Spong(1994)]{dejong1994swinging}
Gerald DeJong and Mark~W Spong.
\newblock Swinging up the acrobot: An example of intelligent control.
\newblock In \emph{Proceedings of 1994 American Control Conference-ACC'94},
  volume~2, pages 2158--2162. IEEE, 1994.

\bibitem[Duan et~al.(2016)Duan, Chen, Houthooft, Schulman, and
  Abbeel]{duan2016benchmarking}
Yan Duan, Xi~Chen, Rein Houthooft, John Schulman, and Pieter Abbeel.
\newblock Benchmarking deep reinforcement learning for continuous control.
\newblock In \emph{International Conference on Machine Learning}, pages
  1329--1338, 2016.

\bibitem[Finn et~al.(2016)Finn, Levine, and Abbeel]{finn2016guided}
Chelsea Finn, Sergey Levine, and Pieter Abbeel.
\newblock Guided cost learning: Deep inverse optimal control via policy
  optimization.
\newblock In \emph{International conference on machine learning}, pages 49--58,
  2016.

\bibitem[Fu et~al.(2017)Fu, Luo, and Levine]{fu2017learning}
Justin Fu, Katie Luo, and Sergey Levine.
\newblock Learning robust rewards with adversarial inverse reinforcement
  learning.
\newblock \emph{arXiv preprint arXiv:1710.11248}, 2017.

\bibitem[Ho and Ermon(2016)]{ho2016generative}
Jonathan Ho and Stefano Ermon.
\newblock Generative adversarial imitation learning.
\newblock In \emph{Advances in neural information processing systems}, pages
  4565--4573, 2016.

\bibitem[Kochenderfer()]{kochenderferjrlf}
Mykel Kochenderfer.
\newblock Jrlf: Java reinforcement learning framework, 2006.
\newblock \emph{URL http://mykel. kochenderfer. com/jrlf}.

\bibitem[Kostrikov(2018)]{pytorchrl}
Ilya Kostrikov.
\newblock Pytorch implementations of reinforcement learning algorithms.
\newblock \url{https://github.com/ikostrikov/pytorch-a2c-ppo-acktr-gail}, 2018.

\bibitem[Mahler and Goldberg(2017)]{mahler2017learning}
Jeffrey Mahler and Ken Goldberg.
\newblock Learning deep policies for robot bin picking by simulating robust
  grasping sequences.
\newblock In \emph{Conference on robot learning}, pages 515--524, 2017.

\bibitem[Moore(1990)]{moore1990efficient}
Andrew~William Moore.
\newblock Efficient memory-based learning for robot control.
\newblock 1990.

\bibitem[Muller et~al.(2006)Muller, Ben, Cosatto, Flepp, and
  Cun]{muller2006off}
Urs Muller, Jan Ben, Eric Cosatto, Beat Flepp, and Yann~L Cun.
\newblock Off-road obstacle avoidance through end-to-end learning.
\newblock In \emph{Advances in neural information processing systems}, pages
  739--746, 2006.

\bibitem[Neumann(2006)]{neumann2006reinforcement}
G~Neumann.
\newblock A reinforcement learning toolbox and rl benchmarks for the control of
  dynamical systems.
\newblock \emph{Dynamical principles for neuroscience and intelligent
  biomimetic devices}, page 113, 2006.

\bibitem[Pomerleau(1989)]{pomerleau1989alvinn}
Dean~A Pomerleau.
\newblock Alvinn: An autonomous land vehicle in a neural network.
\newblock In \emph{Advances in neural information processing systems}, pages
  305--313, 1989.

\bibitem[Schaal(1999)]{schaal1999imitation}
Stefan Schaal.
\newblock Is imitation learning the route to humanoid robots?
\newblock \emph{Trends in cognitive sciences}, 3\penalty0 (6):\penalty0
  233--242, 1999.

\bibitem[Schaul et~al.(2010)Schaul, Bayer, Wierstra, Sun, Felder, Sehnke,
  R{\"u}ckstie{\ss}, and Schmidhuber]{schaul2010pybrain}
Tom Schaul, Justin Bayer, Daan Wierstra, Yi~Sun, Martin Felder, Frank Sehnke,
  Thomas R{\"u}ckstie{\ss}, and J{\"u}rgen Schmidhuber.
\newblock Pybrain.
\newblock \emph{Journal of Machine Learning Research}, 11\penalty0
  (24):\penalty0 743--746, 2010.

\bibitem[Schulman et~al.(2017)Schulman, Wolski, Dhariwal, Radford, and
  Klimov]{schulman2017proximal}
John Schulman, Filip Wolski, Prafulla Dhariwal, Alec Radford, and Oleg Klimov.
\newblock Proximal policy optimization algorithms.
\newblock \emph{arXiv preprint arXiv:1707.06347}, 2017.

\bibitem[Stone et~al.(2005)Stone, Kuhlmann, Taylor, and Liu]{stone2005keepaway}
Peter Stone, Gregory Kuhlmann, Matthew~E Taylor, and Yaxin Liu.
\newblock Keepaway soccer: From machine learning testbed to benchmark.
\newblock In \emph{Robot Soccer World Cup}, pages 93--105. Springer, 2005.

\bibitem[Vinyals et~al.(2017)Vinyals, Ewalds, Bartunov, Georgiev, Vezhnevets,
  Yeo, Makhzani, K{\"u}ttler, Agapiou, Schrittwieser,
  et~al.]{vinyals2017starcraft}
Oriol Vinyals, Timo Ewalds, Sergey Bartunov, Petko Georgiev, Alexander~Sasha
  Vezhnevets, Michelle Yeo, Alireza Makhzani, Heinrich K{\"u}ttler, John
  Agapiou, Julian Schrittwieser, et~al.
\newblock Starcraft ii: A new challenge for reinforcement learning.
\newblock \emph{arXiv preprint arXiv:1708.04782}, 2017.

\bibitem[Wang et~al.(2019)Wang, Devin, Cai, Kr{\"a}henb{\"u}hl, and
  Darrell]{wang2019monocular}
Dequan Wang, Coline Devin, Qi-Zhi Cai, Philipp Kr{\"a}henb{\"u}hl, and Trevor
  Darrell.
\newblock Monocular plan view networks for autonomous driving.
\newblock \emph{arXiv preprint arXiv:1905.06937}, 2019.

\bibitem[Yamaguchi and Ogasawara(2010)]{yamaguchi2010skyai}
Akihiko Yamaguchi and Tsukasa Ogasawara.
\newblock Skyai: Highly modularized reinforcement learning library.
\newblock In \emph{2010 10th IEEE-RAS International Conference on Humanoid
  Robots}, pages 118--123. IEEE, 2010.

\bibitem[Zahavy et~al.(2018)Zahavy, Haroush, Merlis, Mankowitz, and
  Mannor]{zahavy2018learn}
Tom Zahavy, Matan Haroush, Nadav Merlis, Daniel~J Mankowitz, and Shie Mannor.
\newblock Learn what not to learn: Action elimination with deep reinforcement
  learning.
\newblock In \emph{Advances in Neural Information Processing Systems}, pages
  3562--3573, 2018.

\bibitem[Ziebart et~al.(2008)Ziebart, Maas, Bagnell, and
  Dey]{ziebart2008maximum}
Brian~D Ziebart, Andrew~L Maas, J~Andrew Bagnell, and Anind~K Dey.
\newblock Maximum entropy inverse reinforcement learning.
\newblock In \emph{Aaai}, volume~8, pages 1433--1438. Chicago, IL, USA, 2008.

\end{thebibliography}

\end{document}